\documentclass{article}
\usepackage[preprint]{neurips_2024}
\usepackage{amsmath}
\usepackage{diagbox}
\usepackage{wrapfig}
\usepackage[utf8]{inputenc} 
\usepackage[T1]{fontenc}    
\usepackage{hyperref}       
\usepackage{url}            
\usepackage{booktabs}       
\usepackage{amsfonts}       
\usepackage{nicefrac}       
\usepackage{microtype}      
\usepackage{xcolor}         
\usepackage[pdftex]{graphicx} 
\usepackage{colortbl}
\usepackage{multirow}
\usepackage[numbers]{natbib}

\title{CiliaGraph: Enabling Expression-enhanced Hyper-Dimensional Computation in Ultra-Lightweight and One-Shot Graph Classification on Edge}

\author{%
  Yuxi Han \quad\quad\quad  Jihe Wang\quad\quad\quad Danghui Wang \\
  {School of Computer Science, Northwestern Polytechnical University}\\
  \texttt{hyx956@mail.nwpu.edu.cn, wangjihe@nwpu.edu.cn, wangdh@nwpu.edu.cn}
}

\begin{document}

\maketitle
    
\begin{abstract}
Graph Neural Networks (GNNs) are computationally demanding and inefficient when applied to graph classification tasks in resource-constrained edge scenarios due to their inherent process, involving multiple rounds of forward and backward propagation. As a lightweight alternative, Hyper-Dimensional Computing (HDC), which leverages high-dimensional vectors for data encoding and processing, offers a more efficient solution by addressing computational bottleneck. However, current HDC methods primarily focus on static graphs and neglect to effectively capture node attributes and structural information, which leads to poor accuracy. In this work, we propose CiliaGraph, an enhanced expressive yet ultra-lightweight HDC model for graph classification. This model introduces a novel node encoding strategy that preserves relative distance isomorphism for accurate node connection representation. In addition, node distances are utilized as edge weights for information aggregation, and the encoded node attributes and structural information are concatenated to obtain a comprehensive graph representation. Furthermore, we explore the relationship between orthogonality and dimensionality to reduce the dimensions, thereby further enhancing computational efficiency. Compared to the SOTA GNNs, extensive experiments show that CiliaGraph reduces memory usage and accelerates training speed by an average of 292\(\times\)(up to 2341\(\times\)) and 103\(\times\)(up to 313\(\times\)) respectively while maintaining comparable accuracy.
\end{abstract}

\vspace{-0.4cm}
\section{Introduction}
\vspace{-0.2cm}
Graph Neural Networks (GNNs) have significantly advanced the field of graph-based learning, offering powerful models to handle structured data \cite{kipf2017semisupervised,fan2019graph,wu2022graph,wu2020comprehensive,Zhang2018AnED}. However, their reliance on substantial computational and memory resources, coupled with the necessity for multiple training iterations \cite{lecun2015deep}, renders them inefficient, particularly in edge computing scenarios where resources are at a premium. This limitation motivates the exploration of alternative computational frameworks that can offer both efficiency and cost reduction \cite{Zhou2021OptimizingME}.

\begin{wrapfigure}{R}{0.485\textwidth}
  \centering
  \includegraphics[width=0.485\textwidth]{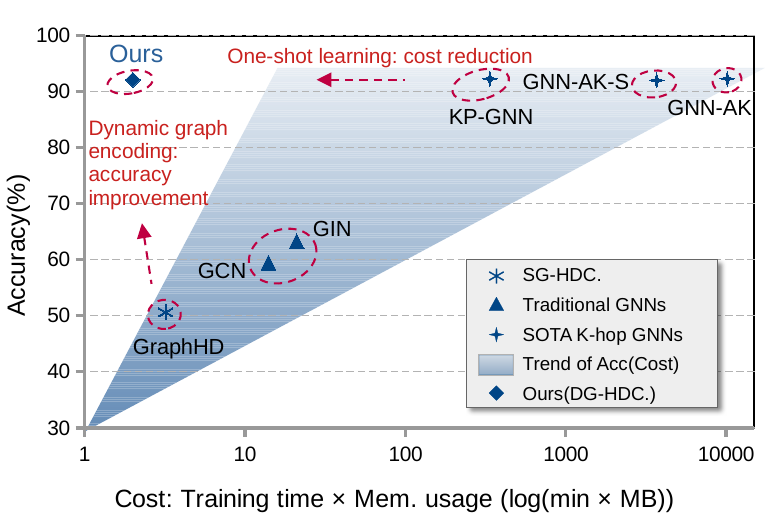}
  \vspace{-0.7cm}
  \caption{Accuracy vs. training cost}
  \vspace{-0.3cm}
  \label{fig:parto}
\end{wrapfigure}

Hyperdimensional Computing (HDC), inspired by the rigorous high-dimensional nature of the human brain's processing capabilities and known for its \textbf{one-shot computational properties}, emerges as a promising resource-friendly and efficient alternative \cite{Kanerva2009HyperdimensionalCA}. HDCs use high-dimensional random vectors (called \emph{hypervectors}) for encoding and processing information, has demonstrated potential in various domains \cite{imani2019quanthd,khaleghi2022generic,kovalev2022vector,kim2020geniehd,li2023hypernode}. While pioneering efforts such as GraphHD \cite{nunes2022graphhd} explore graph classification but focus on static graphs(SG), leading to suboptimal accuracy due to missing key information. As illustrated in Figure ~\ref{fig:parto}, SOTA k-hop GNNs \cite{zhao2021stars,feng2022powerful} have demonstrated remarkable performance in encoding raw features, further enhanced by learnable weights that capture edge importance and connectivity strength through multi-iteration updates. However, SG-HDC suffers from \textbf{hyper-dimensional node encoding distortion} when handling hypervectors, failing to preserve node distance isomorphism. Despite HDC's one-shot learning nature reducing computational overhead compared to GNNs and rendering it a promising framework for resource-constrained edge scenarios \cite{chandrasekaran2022fhdnn,kleyko2022survey}. Yet, the existing SG-HDC exhibits \textbf{asymmetric indiscriminate aggregation}, unable to aggregate all nodes while disregarding edge weights, hindering edge information capture \cite{nunes2022graphhd}. Furthermore, \textbf{node feature loss} during bundling operations undermine HDC's accuracy in obtaining graph-level representations \cite{gilmer2017neural}. Addressing these limitations is crucial to fully leveraging HDC's efficiency while maintaining accuracy on graph classification tasks for edge scenarios.

To tackle these challenges, our work adopts a dynamic graph perspective and proposes a novel HDC-based graph classification algorithm called \textbf{CiliaGraph}. \textbf{First}, CiliaGraph considers inherent node feature and encodes attributes into hypervectors based on their distribution, effectively preserving node distance isomorphism. \textbf{Second}, CiliaGraph uses hypervector similarity distances as edge weights and introduces a transition matrix to smooth the influence of node degrees, facilitating information flow during node aggregation. \textbf{Third}, it preserves the original node features when obtaining the graph-level representation, thus enabling comprehensive graph structure learning. CiliaGraph not only overcomes the static limitations of existing HDC methods but also significantly boosts accuracy on graph classification tasks, approaching the performance of SOTA k-hop GNNs. Our contributions are summarized as follows:

\begin{itemize}
\item We analyze three primary limitations contributing to the poor accuracy of existing HDC methods for graph classification. Based on these insights, we propose CiliaGraph that delivers an ultra-lightweight and efficient solution without compromising accuracy.
\item We introduce a encoding approach that preserves node distance isomorphism and devise an edge weight matrix based on hypervector similarities to capture structural information.
\item We present comprehensive experimental results showing that CiliaGraph is applicable to multiple types of graph datasets, with performance comparable to SOTA GNNs models and significantly better efficiency than GNNs.
\end{itemize}

\vspace{-0.45cm}
\section{Related Works}
\label{gen_inst}
\vspace{-0.2cm}
\textbf{Graph Neural Networks on Edge Devices.} In order to improve the expressiveness of GNNs and overcome the limitations imposed by the Weisfeiler-Lehman (1-WL) test \cite{leman1968reduction}, many new variants have been proposed \cite{zhao2021stars,feng2022powerful,xu2018powerful,kipf2017semisupervised}. These advances enable more comprehensive feature extraction by incorporating a wider range of contextual information from the graph structure. However, these advanced GNN models incur significant overheads in terms of computational cost and memory usage \cite{ding2022sketch}. Memory efficiency issues and OOM problems are often faced on edge platforms. While there have been efforts to compress GNNs to address these issues through strategies such as PCA dimension reduction \cite{jin2020self} and subgraph sampling \cite{zhao2021stars}, these solutions tend to remain computationally intensive.The inherent nature of GNN training using backpropagation imposes additional overheads, exacerbating the challenges of deploying these networks in resource-limited environments \cite{wu2019simplifying,amrouch2022brain}.

\textbf{Hyperdimensional Computing for Graph Data.} In recent years, there have been several research attempts to use HDC to embed graph data \cite{poduval2022graphd,nunes2022graphhd,li2023hypernode,kang2022relhd}. \cite{poduval2022graphd} explores the use of HDC for cognitive graph memory, focusing on efficient information retrieval and memory reconstruction. HyperNode \cite{li2023hypernode}, RelHD \cite{kang2022relhd} are efficient node-level learning models, but are not applicable to graph-level representation learning. GraphHD \cite{nunes2022graphhd} represents an initial benchmark attempt to apply HDC to graph classification tasks, yet it exhibits significant limitations due to its focus on static graphs. GraphHD does not consider node attributes and edge weights, fundamentally lacking a robust mechanism for encoding node attributes. Moreover, GraphHD fails to support the dynamics of message-passing, thereby restricting its ability to capture complex interactive processes within the graph. In terms of edge representation, GraphHD simply binds connected nodes together without considering the variable connection strengths, oversimplifying the graph structure. This approach not only neglects the nuances of the graph's structural topology but also leads to a substantial loss of topological detail when deriving graph-level representations.

\begin{figure}[t]
    \centering \includegraphics[width=1\linewidth]{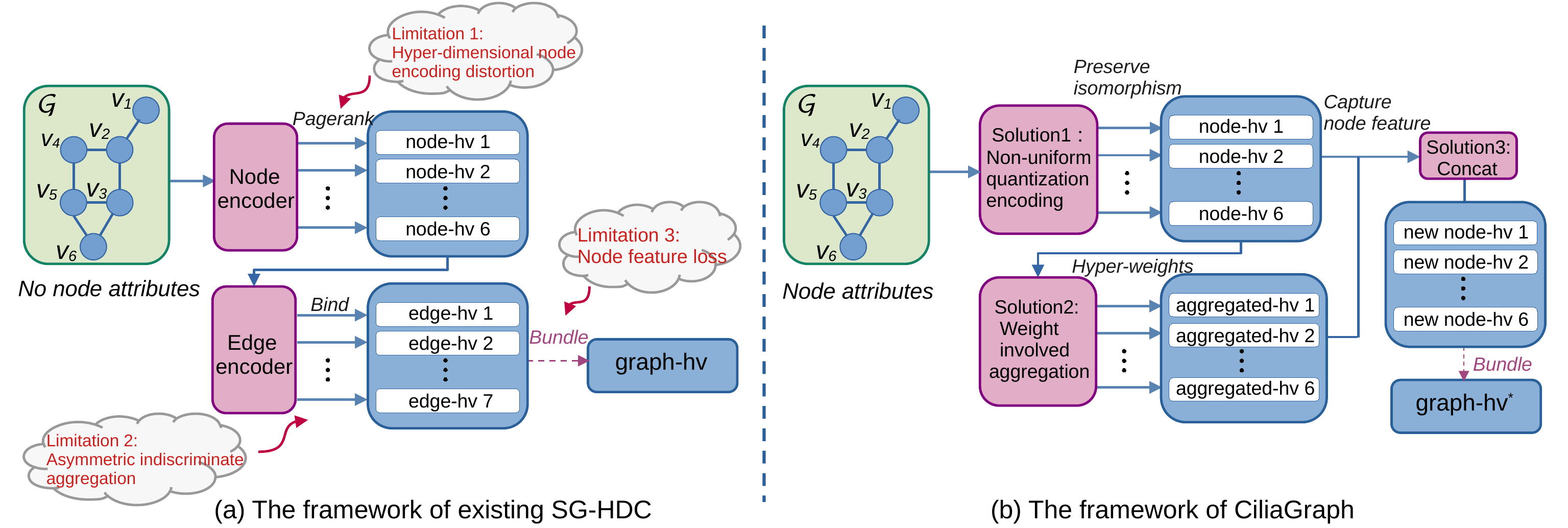}
    \vspace{-0.6cm}
    \caption{Limitations of SG-HDC and our solution}
    \label{fig:hdcmodel}
        \vspace{-0.6cm}
\end{figure}

\section{Limitations of SG-HDC Approaches}
\label{headings}
\vspace{-10pt}
This section outlines three limitations causing low accuracy in existing HDCs and propose our framework accordingly. An overview of the framework is shown in Figure ~\ref{fig:hdcmodel}.

\textbf{Basic Notations.} HDC models employ \emph{D}-dimensional binary hypervectors as the fundamental elements for computation in \(\mathcal{HD}\)(high-dimensional) space. The hypervectors are denoted as \(\mathcal{H}=\{h_1 ,h_2 ,\dots, h_D\}\), where \(h_i \in \{-1, 1\}\). Typically, \(D\) is of the order of thousands \cite{yang2023device}. \(\bigotimes\) represents the binding operation, which combines two hypervectors by performing a bit-wise multiplication, resulting in a new vector. \(\bigoplus\) represents the bundling operation, aggregating a set of hypervectors through bit-wise addition, producing a hypervector that is similar to all operands and represents a set of information \cite{Kanerva2009HyperdimensionalCA,thomas2021theoretical}. \(\delta(\cdot)\) is the similarity function to measure the the distance between vectors in the \(\mathcal{HD}\) space. \(\delta(\cdot)\) generally utilizes the Hamming distance or dot product \cite{hassan2021hyper,yu2022understanding}. All cognitive tasks in HDC are ultimately based on similarity \cite{ge2020classification}. Let each graph \(\mathcal{G} = (\mathcal{V}, \mathcal{E})\) in the graph dataset \(\mathbf{G}\), where \(\mathcal{V}\) is the set of nodes and \(\mathcal{E}\) is the set of edges. \(|\cdot|\) indicates the number of elements in the set. In this work, the focus is on undirected graphs, assuming that both \((u,v)\) and \((v,u)\) belong to \(\mathcal{E}\). 

\textbf{Limitation 1: Hyper-dimensional node encoding distortion.} The encoding method is fundamental to the performance and complexity of HDC and GNN models \cite{aygun2023learning}. In HDC, the encoding process maps attribute vectors \(\mathcal{F}= \{f_1 ,f_2 ,\dots, f_n\mid f_i \in \mathbb{R}\}\) into a \(\mathcal{HD}\) space. Quasi-orthogonality is crucial for unrelated data, while preserving relative distance relationships is essential for correlated data \cite{rahimi2016hyperdimensional,imani2021revisiting}. However, existing encoding methods fail to preserve node distance isomorphism between graph and \(\mathcal{HD}\) spaces when encoding graph nodes with hypervectors, leading to \emph{distortion} and lower accuracy (Section \ref{sec:encoding}). Figure ~\ref{fig:hdcmodel} shows SG-HDC models employs the PageRank \cite{brin1998anatomy} algorithm to allocate hypervectors for each node \(v_i \in \mathcal{V}\), considering only the position and ignoring the intrinsic node attributes. A classic method for incorporating attributes into the encoding process is \emph{record-based encoding} \cite{rahimi2016hyperdimensional,imani2018hierarchical,duan2022hdlock}. This approach quantizes the feature value domain \([f_{min}, f_{max}]\) into \(m\) discrete levels, then assigns a \emph{level hypervector} \(\{\mathcal{L}_1 ,\mathcal{L}_2 ,\dots, \mathcal{L}_m\}\) to each level. All \(f_i\) share these hypervectors, while a set of orthogonal hypervectors distinguishes the indices \(i \in \{1,2,\dots,n\}\) \cite{hassan2021hyper}. \(\mathcal{L}_1\) is randomly generated, and \(\mathcal{L}_i\) is obtained by flipping a fixed \(\frac{D}{2(m-1)}\) bits of \(\mathcal{L}_{i-1}\) \cite{imani2017voicehd}, which limits expressiveness. Inspired by information entropy, a recent approach \cite{nunes2023extension}  allows for variable bit flipping, ensuring hypervector distances are proportional to level differences. However, these methods focus on generating hypervectors based on relationships between quantized discrete levels \(\{1,2,\dots,m\}\), ignoring the data distribution and variance of continuous attributes before quantization. In Section ~\ref{sec:4-1}, a non-uniform quantization encoding is proposed to preserve isomorphism.

\textbf{Limitation 2: Asymmetric indiscriminate aggregation.} Existing SG-HDC model processes neighbor nodes asymmetrically during aggregation. Using graph \(\mathcal{G}\) from Figure ~\ref{fig:hdcmodel} as an example. After getting the node hypervectors \(\mathcal{H}_i\) of \(v_i(i \in \{1,2,\dots,6\})\), SG-HDC directly binds the hypervectors of the connected nodes to represent the edge hypervectors and then bundle them to get the graph-level representation. The common elements are extracted from this process, which can be transformed into:
\begin{equation}
\label{eq:1}
\mathcal{H}_2 \otimes (\underbrace{\mathcal{H}_1 + \mathcal{H}_3 + \mathcal{H}_4}_{\mathcal{H}_2 \: aggregation}) \oplus \mathcal{H}_5 \otimes \underbrace{(\mathcal{H}_3 + \mathcal{H}_4)}_{\mathcal{H}_5 \: aggregation} \oplus \: \mathcal{H}_6 \otimes \underbrace{(\mathcal{H}_3 + \mathcal{H}_5)}_{\mathcal{H}_6 \: aggregation}
\end{equation}
For \(v_2\) , its neighbours  \(v_1\),  \(v_3\) and  \(v_4\) are aggregated as \(\mathcal{H}_2 \otimes (\mathcal{H}_1 + \mathcal{H}_3 + \mathcal{H}_4)\).  However, the relationship between \(v_3\) and \(v_2\) is not similarly considered. GNNs perform the aggregation operation for each node \cite{hamilton2017inductive}, but only for three nodes in SG-HDC. The asymmetry leads to an incomplete flow of information and neglects certain neighbor relationships, thereby affecting the understanding of the overall graph structure. Another significant flaw is that SG-HDC treats all edge relationships indiscriminately during node aggregation and does not distinguish the importance difference of different edges. From Equation ~\eqref{eq:1}, SG-HDC binds the features of \(v_1\),  \(v_3\) and  \(v_4\) to \(v_2\) without employing any weighting mechanism, treating each edge as structurally and semantically equivalent by default. However, in real scenarios, the information carried by different edges can vary significantly and should be given different weights \cite{wind2012weighted,balcilar2021breaking}, which is ignored by SG-HDC and potentially leads to information loss and inaccurate aggregation results. This \textbf{asymmetric indiscriminate aggregation} approach destroys the interactions between nodes and ignores the importance variations of different edges \cite{corso2020principal}, which constrains SG-HDC's ability to represent graph data. In Section ~\ref{sec:4-2}, the distance of hypervectors is involved to distinguish the connection strength in graph, and node degrees are incorporated to achieve symmetric aggregation.

\textbf{Limitation 3: Node feature loss.}
After aggregating node hypervectors, SG-HDC bundles all these results to form the final graph-level representation \(\mathcal{H}_\mathcal{G}\). However, SG-HDC discards the original features of each node during the bundling process, retaining only the aggregated outcomes. As a result, node information is lost in the final graph representation, only interactions with neighbors are preserved. The drawback of this approach is significant: original node features contain unique attribute information crucial for a comprehensive description of the graph \cite{xu2018powerful,velickovic2017graph}. The loss of these unique information degrades the quality of the graph representation. In Section ~\ref{sec:4-3}, CiliaGraph's concatenation operation is proposed following GNNs-manner that compensate node feature for the comprehensive description.   

\vspace{-0.3cm}
\section{CiliaGraph Architecture}
\label{others}
\vspace{-0.2cm}

\subsection{Non-uniform dynamic encoding}
\label{sec:4-1}
\vspace{-0.2cm}
\textbf{Non-uniform dynamic initialization.}
Typically, uniform quantization \cite{han2015deep} divides values into equal intervals, often resulting in sparse data representation and inaccuracies (Section ~\ref{sec:level and dimension}). Conversely a non-uniform clustering quantization is implemented using the K-means \cite{han2015deep} algorithm. This method categorizes node attribute values into \(m(m>2)\) clusters based on their actual distribution, ensuring approximately equal data volume within each cluster. Each cluster corresponds to a quantization center \(\{\mu_1 ,\mu_2 ,\dots, \mu_m\}\), representing the quantized \(m\) levels. 

\begin{wrapfigure}{R}{0.28\textwidth}
  \centering
  \includegraphics[width=0.28\textwidth]{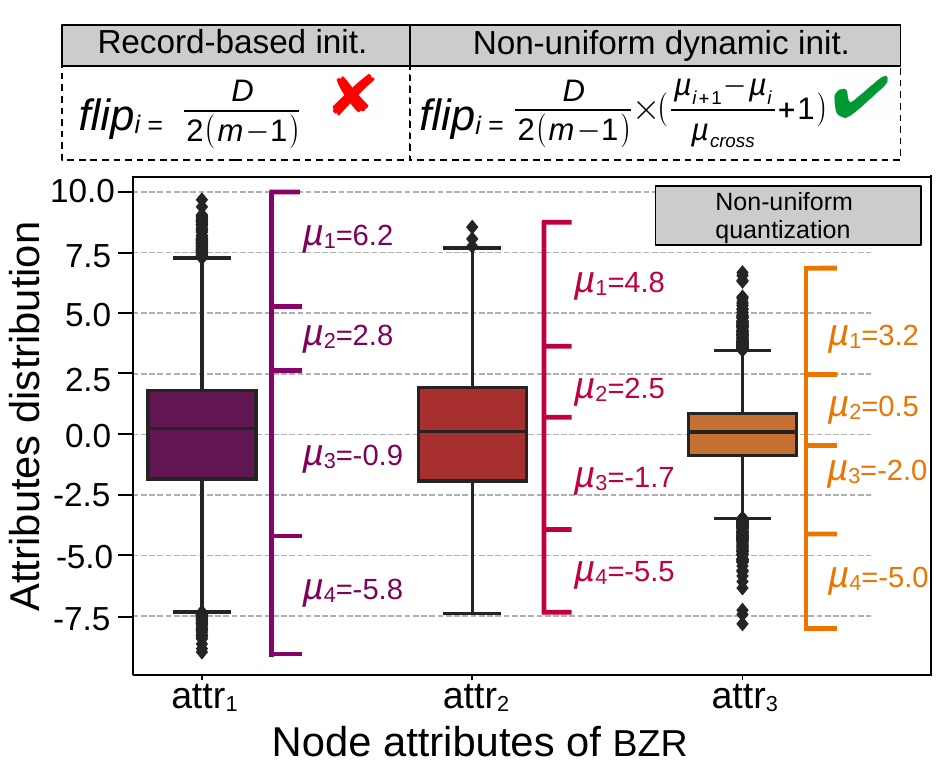}
  \caption{Variable attribute distributions drive individual dynamic encoding for these attributes.}
  \label{fig:group_distribution}
  \vspace{-0.4cm}
\end{wrapfigure}

Our initialization method randomly generates a \(D\)-dimensional hypervector, \(\mathcal{L}_1\), from \(\{1, -1\}\), corresponding to the \(\mu_1\). Figure \ref{fig:group_distribution} shows the attribute values distribution for the BZR dataset \cite{dataset-bzr}. Each attribute distribution is uneven and the distances between cluster centers are variable. Traditional methods flip a fixed number of bits, ignoring original data correlations. Our approach dynamically adjusts the bit-flipping proportion based on the differences between non-uniform clusters, with the number of bits flipped determined by the Euclidean distances between adjacent cluster centers: 
\begin{equation}
\label{eq:flip}
flip_{i-1} = \frac{D}{2(m-1)}\times\left(\frac{\mu_i -\mu_{i-1}}{\mu_{cross}} + 1\right) \quad \text{for} \quad 1<i\le m
\end{equation}
where \(\mu_{cross} = \mu_m -\mu_1\) controls the overall span of the clustering centers and determines the flip ratio. These bits are flipped once and remain fixed. When \(\mathcal{L}_m\) is generated, \(\mathcal{L}_m\) flips \(\frac{D(m)}{2(m-1)}\) bits compared to \(\mathcal{L}_1\), ensuring quasi-orthogonal between \(\mathcal{L}_1\) and \(\mathcal{L}_m\). 

\colorbox{cyan!15}{%
  \parbox{0.985\textwidth}{%
    \textbf{Proposition 1.} \emph{Assume a set of hypervectors \(\mathbf{L} = \{\mathcal{L}_1, \mathcal{L}_2,\dots, \mathcal{L}_m\}\)are obtained through non-uniform initialization. For any \(j>i \in \{1,2,\dots,m\}\), the expectation of the distance between \(\mathcal{L}_i\) and \(\mathcal{L}_j\), \(\mathbb{E}[\delta(\mathcal{L}_i,\mathcal{L}_j)]\), is proportional to both the quantization level \(j-i\) and the difference in cluster centers \(\mu_j-\mu_i\).}
  }%
}

This initialization method incorporates the original data distribution into the flipping process, enhancing the reliability and expressive power of hypervectors, and improving the representation of continuous attributes in \(\mathcal{HD}\) spaces. Compared to the traditional method of precisely orthogonal flipping \cite{imani2021revisiting}, quasi-orthogonality enriches hypervector representation and facilitates the exploration of lower-dimensional hypervector spaces(see in Section ~\ref{low-dimension}) \cite{nunes2023extension}.

\textbf{Nodes encoding.} The record-based method works when all attributes \(f_i\) share the same set of level hypervectors \cite{hassan2021hyper}. However, Figure ~\ref{fig:group_distribution} shows that variable value distributions of graph node attributes make a single set of level hypervectors unsuitable. Our approach generates a set of internally correlated \(\mathcal{L}^{f_i}=\{\mathcal{L}^{f_i}_{1}\allowbreak,\mathcal{L}^{f_i}_{2}\allowbreak,\dots\allowbreak,\mathcal{L}^{f_i}_{m}\allowbreak\}\) for each attribute category \(f_i\), \(i \in \{1,2,\dots,n\}\), ensuring the sets \(\mathcal{L}^{f_i}\) among different attributes are unrelated and quasi-orthogonal. This design provides distinct representations for different attributes and satisfies quasi-orthogonality between hypervectors involved in encoding operations \cite{yang2023device,Kanerva2009HyperdimensionalCA}. After obtaining all \(\mathcal{L}^{f_i}\), a node \(u\) in a graph \(\mathcal{G} = (\mathcal{V}, \mathcal{E})\) is encoded, which has a quantized attribute vector \(\{m_{u1}, m_{u2},\dots,m_{un}\}\): \(\mathcal{H}_u = \bigoplus^{n}_{i=1} \mathcal{L}^{f_i}_{ui}\), where \(ui\) denotes the quantization level on the \(i\)-th attribute. Encoding all \(u \in \mathcal{V}\) yields the node hypervector matrix \(\mathbf{H}_{\mathcal{V}} \in 
\mathbb{R}^{\mathcal{|V|} \times D}\). Our initialization and node encoding methods comprehensively capture the distribution of the original continuous data, ensuring that the expectation of the distance between hypervectors relates to the quantization centers, which preserves the isomorphism between node distances in graph and \(\mathcal{HD}\) spaces. Since bits are flipped only once, node differences are effectively reflected in the number of differing bits in the non-binary hypervectors, which facilitates further processing (Section ~\ref{sec:4-2}).

\vspace{-0.3cm}
\subsection{Weight involved aggregation}
\label{sec:4-2}
\vspace{-0.15cm}
GNNs aggregate each node through a neighborhood aggregation strategy \cite{ijcai2020p181}, introducing MLP or learnable parameters in each layer \cite{nikolentzos2019message}, which are updated through multiple rounds of back-propagation to better capture latent node relationships and edge weights. Unlike GNNs, HDC lacks iterative parameter updates, making it lightweight and computationally simple \cite{kim2023efficient}. However, HDC struggles to capture latent connections and edge strengths between nodes accurately. In Section ~\ref{sec:4-1}, a novel encoding technique is introduced that preserves node distance isomorphism in the \(\mathcal{HD}\) space. To represent node connections based on their distances, a \emph{similarity weight matrix} \(\mathbf{W}_\mathcal{G} \in \mathbb{R}^{\mathcal{|V|} \times \mathcal{|V|}}\) is proposed for graph \(\mathcal{G}\), leveraging hypervector distances to effectively model relationships between connected nodes. Since node differences are reflected in the varying bits of hypervectors, the Hamming distance is used to compute their similarity. 

\begin{figure}
    \centering
    \includegraphics[width=1\linewidth]{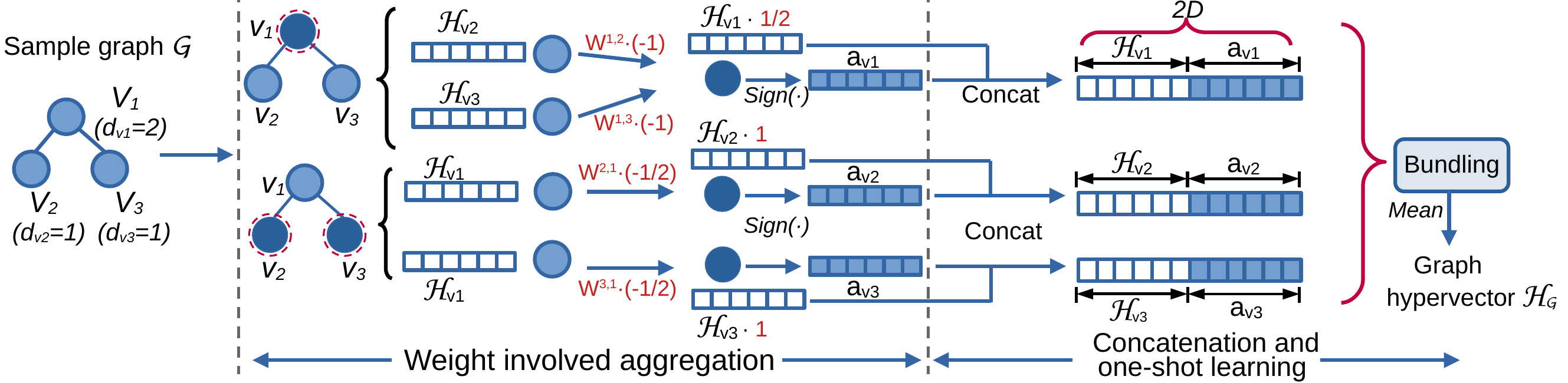}
    \caption{The processes of weight involved aggregation and one-shot learning: symmetric aggregation with weights, followed by concatenation to preserve comprehensive features.}
    \vspace{-0.35cm}
    \label{fig:4-2}
\end{figure}

In addition, a transition matrix \(\mathbf{T}_\mathcal{G} \in 
\mathbb{R}^{\mathcal{|V|} \times \mathcal{|V|}}\) is presented to incorporate nodal degree information, adjusting weights to enhance the graph's representation. This transition matrix models the probability and intensity of information propagation between nodes, enriching edge semantics. Assuming that the degree of a node is denoted by \(d\). To smooth out degree effects, weights are normalized by the inverse of the degree: \(\mathbf{T}_\mathcal{G}^{u,v}=-\frac{1}{d_v}\) for \((u,v)\in \mathcal{E}\), and \(\mathbf{T}_\mathcal{G}^{u,u}=\frac{1}{d_u}\) when introducing self-loops. The transition matrix ensures that neighboring nodes' contributions are weighted proportionally based on their degrees. The normalized \(\mathbf{W}_\mathcal{G}\) and \(\mathbf{T}_\mathcal{G}\) form the \textbf{Hyper-weight matrix} \(\mathbf{P}_\mathcal{G} = \mathbf{W}_\mathcal{G} \times \mathbf{T}_\mathcal{G}\). Combining these matrices addresses the issue of indiscriminate edge treatment and incorporates nodal degree influence,  enhancing the expressive power and facilitating a comprehensive understanding of the dynamics within the graph, yielding a more enriched representation of its structure. The aggregation operation is as follows (Figure ~\ref{fig:4-2}):
\begin{equation}
a_u = Sign\left(\bigoplus_{v \in \mathcal{N}_{u}^\ast}\left({\mathbf{P}_{\mathcal{G}}^{u,v}} \times \mathcal{H}_{v}\right)\right) \quad \text{with} \quad \mathcal{N}_{u}^\ast = \mathcal{N}_{u} \cup \{u\}
\end{equation}
where \(\mathcal{N}_{u}\) represents the set of all nodes adjacent to \(u\). The sign function \cite{bric} maps the aggregation result to +1 or -1 values, with \(a_u\) capturing the aggregated messages from all nodes adjacent to \(u\) (including the self-loop), reflecting the topology and the weighted interactions between nodes. 

\vspace{-0.3cm}
\subsection{One-shot learning of node attributes and graph topology}
\label{sec:4-3}
\vspace{-0.15cm}
To address node feature loss during bundling, we concatenate \(\mathcal{H}_{u}\) with \(a_u\) (Figure ~\ref{fig:4-2}). This preserves both node properties and aggregated topological information, enhancing the model's expressiveness and allowing for a more comprehensive representation. The concatenation is expressed as:
\vspace{-4pt}
\begin{equation}
\begin{aligned}
\Tilde{\mathcal{H}}_{u} &= CONCAT(\mathcal{H}_{u},a_u) \\
&= CONCAT\left(\bigoplus^{n}_{i=1} \mathcal{L}^{f_i}_{ui}, Sign\left(\bigoplus_{v \in \mathcal{N}_{u}^\ast}\left({\mathbf{P}_{\mathcal{G}}^{uv}} \times \mathcal{H}_{v}\right)\right)\right)
\quad \text{with} \quad \mathcal{N}_{u}^\ast = \mathcal{N}_{u} \cup \{u\}
\end{aligned}
\end{equation}
\vspace{-10pt}

The result of concatenating \(\Tilde{\mathcal{H}}_{u}\) is analogous to the output of a neural network layer in GNNs. Through this one-shot efficient aggregation method, our model not only retains the essential characteristics of the graph's nodes but also enriches the representation with detailed topological information. The enhanced feature matrix \(\Tilde{\mathbf{H}}_{\mathcal{V}}\) serves as a robust foundation for subsequent graph analysis tasks.

\textbf{CiliaGraph learning}. After the aggregation and concatenation process in CiliaGraph, the graph \(\mathcal{G}\) is processed by first bundling the hypervectors of all nodes in \(\Tilde{\mathbf{H}}_{\mathcal{V}}\), then applying a mean operation to obtain the graph-level hyperdimensional representation \(\mathcal{H}_\mathcal{G}\). Representations of all graphs of the same class are then bundled together, resulting in the model's final required class prototypes \(\mathcal{C}=\{\bigoplus_{\forall\: \mathcal{G}\in class\: i} \mathcal{H}_\mathcal{G} | i\in 1,2,\dots,K\}\).

\textbf{CiliaGraph inference}. During the inference phase, a sample graph undergoes the same process and gets the query \(\mathcal{H}_\mathcal{G}\). The distances between query \(\mathcal{H}_\mathcal{G}\) and all \(\mathcal{C}_i\) are calculated, with the closest one is returned as the prediction. Both \(\mathcal{C}_i \in \mathbb{R}^{D} \) and
\(\mathcal{H}_\mathcal{G} \in \mathbb{R}^{2D}\) are non-integer. Therefore, the cosine function is used to calculate the similarity distance. For efficiency, the L2 norm of each class prototype is computed to transform it into a unit vector \(\mathcal{C}^{norm}_i\), simplifying the similarity calculation \cite{li2023hypernode}:

\colorbox{cyan!15}{%
  \parbox{0.985\textwidth}{%
    \textbf{Theorem 1.} \emph{With the class prototypes \(\mathcal{C}_i\) and their normalized forms \(\mathcal{C}^{norm}_i\), the cosine similarity can be equivalently projected to a dot product. Formally, \(Cos(\mathcal{C}_i,\mathcal{H}_\mathcal{G}) \equiv Dot(\mathcal{C}^{norm}_i,\mathcal{H}_\mathcal{G})\).}
  }%
}

Utilizing the dot product instead of the cosine function efficiently calculates relative distances, simplifying the process without compromising model prediction accuracy.

\vspace{-0.3cm}
\subsection{A discussion: low-dimensional hypervector spaces under quasi-orthogonality}
\label{low-dimension}
\vspace{-0.3cm}
GraphHD relies on high-dimensional hypervectors to enhance accuracy due to its limited ability to fully capture and represent graph data. In contrast, our model effectively preserves isomorphism and node relationship strength while incorporating structural and topological information, achieving accuracy without high dimensions. Drawing from \cite{Yan_2023} and our unique encoding approach, the minimal dimensions required for the framework are investigated, exploring the relationship between quasi-orthogonality and dimensionality.

\colorbox{cyan!15}{%
  \parbox{0.985\textwidth}{%
   \textbf{Definition 1.} \emph{(\(\epsilon\)-quasi-orthogonality). For two unit vectors x and y, if \(|x \cdot y|\le \epsilon\), then they are \(\epsilon\)-quasi-orthogonal} \cite{kainen2020quasiorthogonal}.
  }%
}

where \(\epsilon \in [0,1)\) represents the angle deviation from precisely orthogonality. Next, the vectors are extended into a \(D\)-dimensional bipolar vectors space \(\{\pm1\}^D\). 

\colorbox{cyan!15}{%
  \parbox{0.985\textwidth}{%
    \textbf{Definition 2.} \emph{Assume the dimension is D,the lower bound for the number of \(\epsilon\) -quasi-orthogonal hypervectors that can be found in \(\{\pm1\}^D\) is: \(dim_{\epsilon}(D) \ge e^{D\epsilon^2 / 2}\)} \cite{Yan_2023,kainen2020quasiorthogonal}.
  }%
}

According to Definition 2, as the spatial dimension \(D\) or the quasi-orthogonality threshold \(\epsilon\) increases, the number of quasi-orthogonal vectors can be accommodated grows exponentially. Building on the non-uniform dynamic initialization, a specific value for \(\epsilon = \frac{2}{(m-1)}\) is deduced, consequently deriving the relationship between the number of \(\epsilon\)-quasi-orthogonal hypervectors and the minimum required dimensions.

\colorbox{cyan!15}{%
  \parbox{0.985\textwidth}{%
    \textbf{Proposition 2.} \emph{If n and m represent the number of node attributes and quantization levels separately, then following the generation of n groups of level hypervectors through the Non-uniform initialization method, our quasi-orthogonality threshold \(\epsilon\) is set to \(\frac{2}{(m-1)}\). Consequently, we anticipate a lower bound for \(dim_{\epsilon}(D)\) to be \(2n\). Therefore, according to Definition 2, to ensure at least \(2n\) quasi-orthogonal hypervectors, the minimum dimension D of our hypervectors should satisfy: \(D \approx \frac{(m-1)^2}{2}\ln{2n}\).}
  }%
}

Proposition 2 establishes the relationship among \(D\), \(m\) and \(n\); with \(n\) fixed for a graph dataset, \(D\) is directly proportional to \(m\). Once \(m\) is determined, it facilitates the exploration of achieving quasi-orthogonality with fewer dimensions \(D\), thus reducing computational load while maintaining sufficient model performance.

\vspace{-0.4cm}
\section{Experiments}
\label{experiment}
\vspace{-0.3cm}
In this section, we perform a series of experiments to validate the effective expression power and efficiency of CiliaGraph. We: 1) validate the effectiveness of CiliaGraph's non-uniform quantization, determining the optimal quantization levels and minimal dimensional requirements; 2) demonstrate CiliaGraph's expressive capabilities on four real-world datasets, highlighting its significant advantages in computational overhead and efficiency; 3) establishe five comparative frameworks to validate the efficacy of CiliaGraph's encoding methods; 4) confirm the effectiveness of the Hyper-weights matrix; 5) analyze the balance between accuracy and dimension. The CiliaGraph and GraphHD framework are both implemented using the Torchhd package \cite{torchhd}. All experiments are performed on AMD EPYC CPUs with 96 cores, and two NVIDIA GeForce RTX 4090 GPUs. The seven real-world datasets selected from Tudataset \cite{morris2020tudataset} cover multiple domains, featuring varying complexities in graph structures and node attributes. 

\vspace{-0.4cm}
\subsection{Quantization levels and minimum dimensions in CiliaGraph} 
\label{sec:level and dimension}
\begin{figure}[t]
    \centering
    \begin{minipage}[t]{0.48\textwidth}
        \centering
        \includegraphics[width=\textwidth]{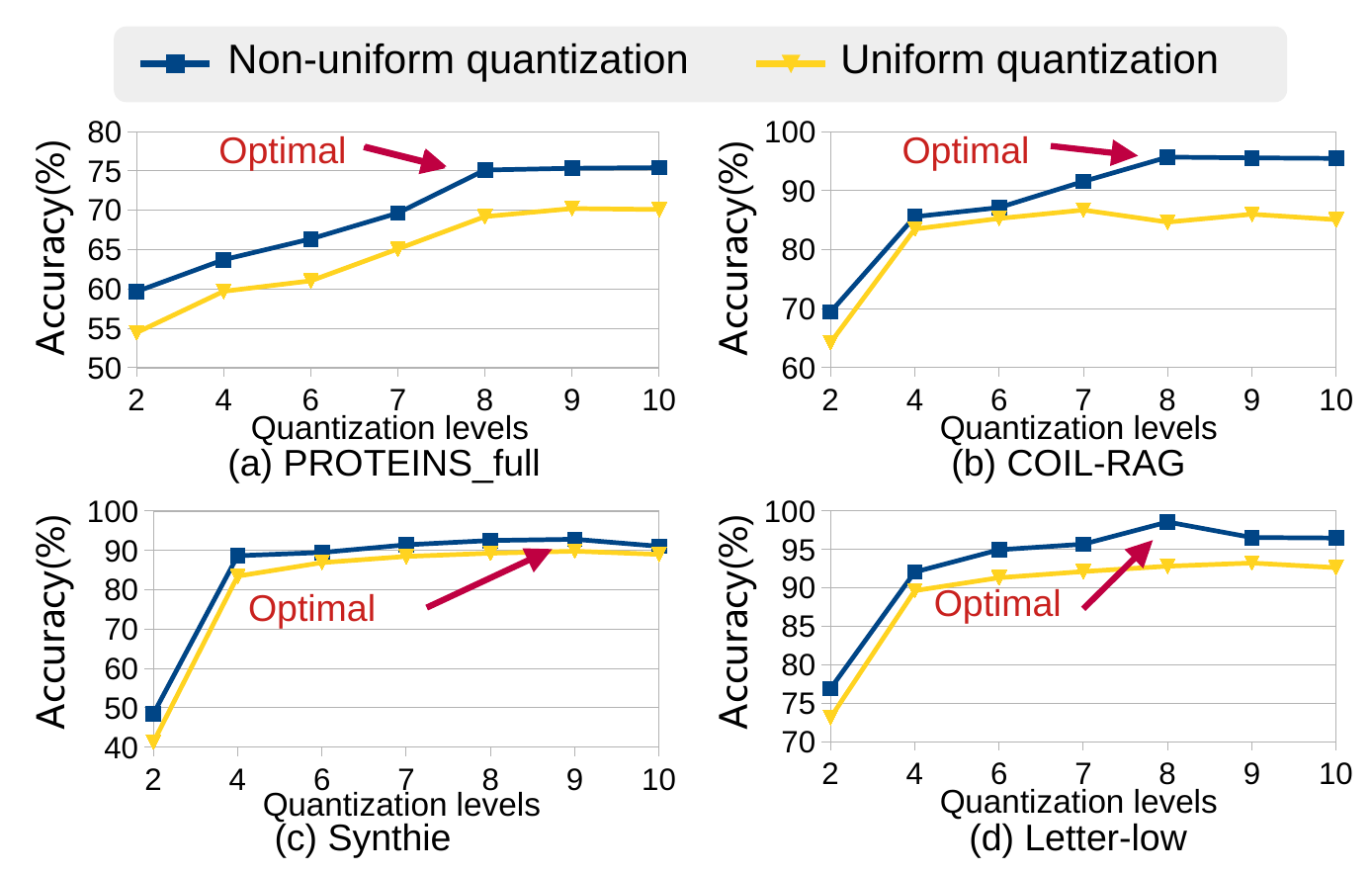}
        \caption{Optimal quantization levels.}
        \label{fig:quantization-levels}
    \end{minipage}
    \hfill 
    \begin{minipage}[t]{0.51\textwidth}
        \centering
        \includegraphics[width=\textwidth]{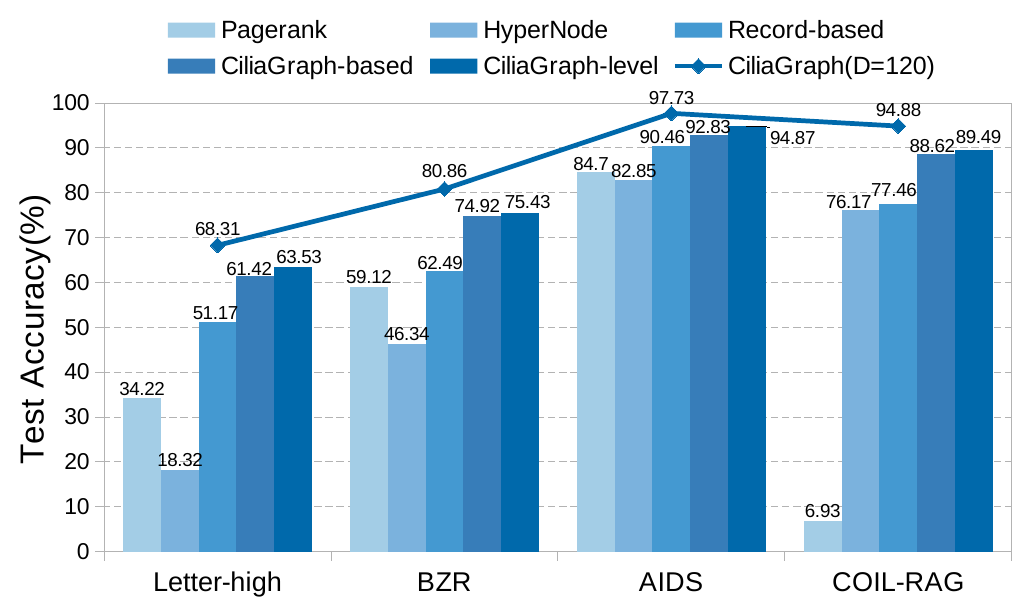}
        \caption{Accuracy of different encoding methods.}
        \label{fig:encode-method}
    \end{minipage}
\vspace{-15pt}
\end{figure}

\vspace{-0.2cm}
The choice of quantization levels significantly affects data clustering and model accuracy. According to the formula provided in Proposition 2, the quantization level also determines the minimum required dimension. Thus, \(D=10,000\) is initially set to identify the optimal quantization levels. Figure ~\ref{fig:quantization-levels} illustrates the results for four datasets, with the yellow line indicating uniform quantization. Most datasets achieve the highest and most stable accuracy at \(\boldsymbol{m=8}\). Subsequently, the formula changes to \(D \approx \frac{(8-1)^2}{2}\ln{2n}\). Therefore, even for the COIL-RAG dataset \cite{morris2020tudataset,dataset-letter}, which has the largest number of node attributes at \(n=64\), the minimum required dimension is \(D \approx \frac{(8-1)^2}{2}\ln{(2\times64)} \approx 118.87\) bits. This substantial reduction in dimension lowers both computational and storage requirements, enhancing the model's efficiency. For convenience in further experiments, \(\boldsymbol{D=120}\) bits will be standardized to evaluate the effectiveness of CiliaGraph. 

\begin{table}[h]
\centering
\caption{Memory, training time, and test accuracy of seven models. The top three are highlighted by \textbf{\textcolor{red}{First}}, \textbf{\textcolor{blue}{Second}}, \textbf{{Third}}.}
\label{tab:methods_comparison}
\resizebox{11cm}{!}{
\small
\begin{tabular}{@{}lccccccccc@{}}
\toprule
\diagbox{Dataset}{Methods} & \parbox[t]{2cm}{\centering GraphHD \\(CPU/GPU)} & GIN & GCN & GIN-AK & GIN-AK-S & KP-GIN & \parbox[t]{2cm}{\centering \textbf{Ours} \\(CPU/GPU) } \\
\midrule
\multicolumn{8}{c}{\textbf{Memory Usage (MB)\(\downarrow\)}} \\
PROTEINS\_full & 120 & 95 & 80 & 3088 & 541 & 285 & \textbf{\textcolor{red}{7}} \\
COIL-RAG & \textbf{\textcolor{red}{6}} & 41 & 30 & 92 & 50 & 62 & \textbf{\textcolor{blue}{8}} \\
Synthie & 55 & 150 & 118  & 9364 & 1414 & 672 & \textbf{\textcolor{red}{4}} \\
Letter-low & 9  & 30  & 22  & 137 & 84 & 76 & \textbf{\textcolor{red}{5}} \\
\multicolumn{8}{c}{\textbf{Training Time (s)\(\downarrow\)}} \\
PROTEINS\_full & 3.2 / 7.6 & 17.2 & 15.5 & 58.8 & 238.2 & 58.7 & \textbf{\textcolor{red}{1.3}} / 1.4 \\
COIL-RAG & 4.1 / 14.4 & 42.4 & 35.6 & 135.8 & 492.3 & 102.6 & \textbf{\textcolor{red}{3.2}} / 4.5 \\
Synthie & 1.8 / 3.1 & 8.5 & 7.4 & 65.2 & 156.7 & 30.1 & \textbf{\textcolor{red}{0.5}} / 0.6 \\
Letter-low & 5.2 / 30.0 & 23.6 & 22.3 & 74.5 & 303.9 & 60.3 & \textbf{\textcolor{red}{1.9}}/ 2.6 \\
\multicolumn{8}{c}{\textbf{Test Accuracy (\%)\(\uparrow\)}} \\
PROTEINS\_full & 64.02 & 68.23 & \textbf{{70.72}} & 67.86 & 65.78 & \textbf{\textcolor{blue}{73.13}} & \textbf{\textcolor{red}{73.96}} \\
COIL-RAG & 6.79 & 90.75 & 89.41 & \textbf{\textcolor{blue}{96.51}} & \textbf{\textcolor{red}{98.36}} & \textbf{{95.01}} & 91.88 \\
Synthie & 50.63 & 63.31 & 59.50 & \textbf{\textcolor{red}{94.15}} & 92.03 & \textbf{92.31} & \textbf{\textcolor{blue}{92.67}} \\
Letter-low & 50.81 & 94.02 & 85.66 & \textbf{\textcolor{red}{98.47}} & 96.28 & \textbf{{96.44}} & \textbf{\textcolor{blue}{97.76}} \\
\bottomrule
\end{tabular}
}
\vspace{-0.4cm}
\end{table}

\vspace{-0.3cm}
\subsection{Performance and efficiency of graph classification tasks} 
\label{sec:5-2}
\vspace{-0.2cm}
For the baseline models, we choose: 1) SG-HDC: GraphHD; 2) Common GNNs: GIN \cite{xu2018powerful}, GCN \cite{kipf2017semisupervised}; 3) SOTA k-hop GNNs: GIN-AK+, GIN-AK+-S \cite{zhao2021stars}, and KPGIN \cite{feng2022powerful}. where GIN-AK+-S serves as a sampling model for GIN-AK+ that can improve resource overhead and training efficiency. All GNNs are run on GPUs, while HDCs are run on both CPUs and GPUs to demonstrate their applicability in resource-constrained scenarios.

The results shown in Table ~\ref{tab:methods_comparison} indicate that CiliaGraph achieves notable accuracy across various datasets and reaches the highest accuracy on the PROTEINS\_full dataset \cite{dataset:proteins}. CiliaGraph achieves \(85.09\%\) higher accuracy than GraphHD on COIL-RAG, demonstrating its advantage in considering intrinsic node attributes. However, its accuracy is \(6.48\%\) lower than GNNs, likely due to the lack of edge attribute processing. Compared to SOTA k-hop GNNs, CiliaGraph reduces memory usage by an average of \(292\times\) and speeds up training by \(103\times\). Particularly on the Synthie dataset \cite{dataset:synthie}, CiliaGraph's performance trails the best GNN model by only \(1.48\%\), while offering a \(2341\times\) reduction in memory usage and a \(313\times\) faster training time.

Additionally, we demonstrate that utilizing GPUs does not improve the performance of HDC models. This is because HDC models require minimal computational resources and operates efficiently without the need for additional decive. The main computational cost arises from data transfers between devices, which becomes a bottleneck and negates the benefits of HDC's high performance. In contrast to GraphHD (\(+234.45\%)\) in time, CiliaGraph \((+26.275\%)\) exhibits smaller performance variations on GPUs due to its memory-friendly low-dimensional design, further affirming its viability and practicality in edge computing scenarios.

\vspace{-20pt}
\begin{wrapfigure}{R}{0.35\textwidth}
  \centering  \includegraphics[width=0.35\textwidth]{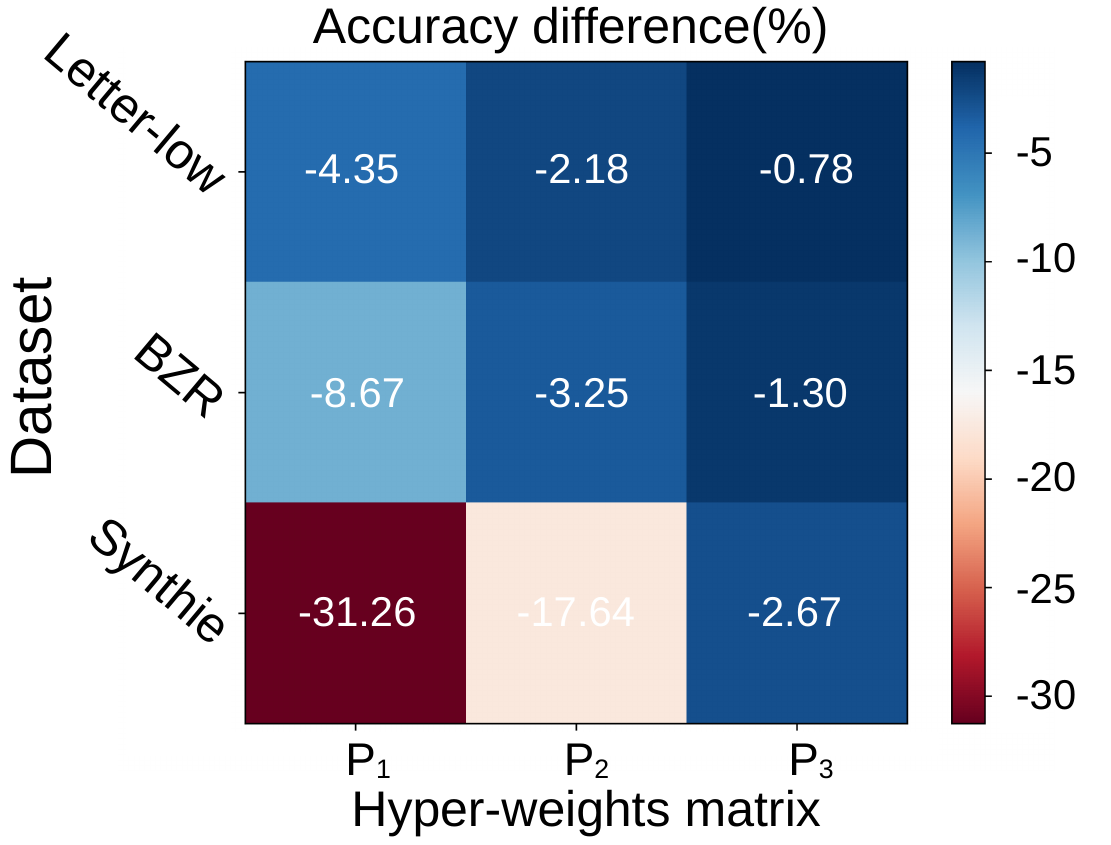}
  \caption{Accuracy loss w/o similarity and transition matrices.}
  \vspace{-10pt}
  \label{fig:laplacian}
\end{wrapfigure}

\vspace{10pt}
\subsection{Expressiveness of the CiliaGraph encoding}
\label{sec:encoding}
\vspace{-0.3cm}
To assess the expressive power of the CiliaGraph node encoding method, we compare it against five alternative frameworks: 1) \textit{PageRank};  2)\textit{HyperNode}; 3)\textit{Record-based}; 4)\textit{CiliaGraph-based}; and 5)\textit{CiliaGraph-level}. All these five frameworks use the dimension \(D=10000\). 

As shown in Figure ~\ref{fig:encode-method}, CiliaGraph achieves optimal performance across all tested datasets, with an average accuracy \(15\%\) higher than the \textit{Record-based} and \(39\%\) higher than the \textit{Pagerank}. This significant improvement underscores the effectiveness of our model and encoding methodology. 

\vspace{-0.3cm}
\subsection{Assessing the efficacy of the Hyper-weights matrix}
\vspace{-0.2cm}
To validate the effectiveness of the more precise similarity weights and transition matrix introduced in CiliaGraph, we compare CiliaGraph with 1) \(\mathbf{P}_1\): \emph{w/o similarity weights matirx and transition matrix}; 2) \(\mathbf{P}_2\): \emph{w/o similarity weights matrix}; and 3) \(\mathbf{P}_3\): \emph{w/o transition matrix}\footnote{w/o means changing the value to 1 while preserving the numerical sign.}. The accuracy differences of these frameworks are shown in Figure ~\ref{fig:laplacian}.
 
It is observed that the performance of all three frameworks progressively deteriorates, particularly for \(\mathbf{P}_1\) and \(\mathbf{P}_2\). This accentuates the inadequacies of basic edge encoding techniques which overlook the significance of connection strengths. The results of \(\mathbf{P}_3\) demonstrate that integrating transition matrices with similarity weight matrices can enhance the model's expressiveness, offering a more refined depiction of inter-node connections. The combination of the weights matrix, offering weights based on similarity, and the transition matrix, adjusting weights in accordance with node transition probabilities, presents a more nuanced, dynamic weighting mechanism that takes into account both feature similarity and interaction intensity among nodes.

\begin{wrapfigure}{R}{0.5\textwidth}
  \centering
  \includegraphics[width=0.5\textwidth]{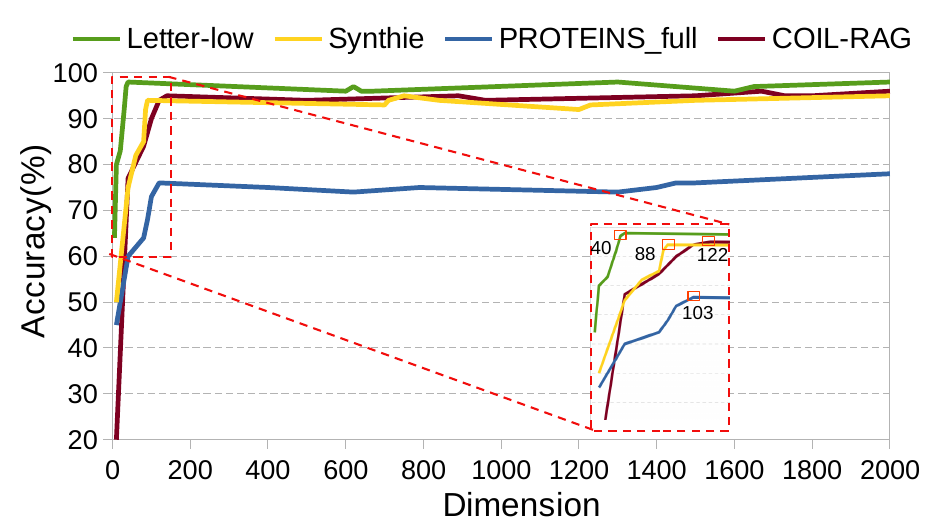}
  \caption{Picking optimal dimension: dimensions vs accuracy.}
  \label{fig:dimension}
  \vspace{-0.4cm}
\end{wrapfigure}

\vspace{-0.4cm}
\subsection{Analysis of optimal minimum dimension}
\label{sec:5-5}
\vspace{-0.2cm}
The minimum dimensions required for each dataset are calculated based on a previously established formula, which considers the complexity of the data. As shown in Figure ~\ref{fig:dimension}, our experiments span from dimensions below the calculated minimum \((D=5)\) to settings well above it \((D=2000)\). Results indicate that accuracies are considerably lower below the minimum dimensions, confirming the theoretical predictions. Upon reaching or exceeding the minimum dimensions, accuracies improve significantly but soon plateau, suggesting that additional dimensions beyond the necessary minimum yield diminishing returns. This behavior underscores the importance of selecting an appropriate dimension for efficient graph classification in HDC, aligning closely with our theoretical framework on the optimal dimension for such tasks. In addition, the generated initial hypervectors exhibit not only intra-group correlation but also inter-group quasi-orthogonality among different sets, as described in Section ~\ref{sec:4-1}.

\vspace{-0.4cm}
\section{Conclusion}
\label{sec:conclusion}
\vspace{-0.3cm}
In this paper, we presented CiliaGraph, a highly efficient and ultra-lightweight graph classification framework tailored for edge computing. CiliaGraph is adept at preserving the isomorphism of node distances, capturing an extensive range of structural information. Our experiments confirm that CiliaGraph achieves commendable, and at times superior, accuracy with minimal time and computational resource expenditure, pushing the boundaries of the SOTA methods. This framework sets a new benchmark for rapid deployment and effective graph analysis in resource-constrained environments.

\bibliographystyle{unsrt}

\begin{thebibliography}{10}

\bibitem{kipf2017semisupervised}
Thomas~N. Kipf and Max Welling.
\newblock Semi-supervised classification with graph convolutional networks, 2017.

\bibitem{fan2019graph}
Wenqi Fan, Yao Ma, Qing Li, Yuan He, Eric Zhao, Jiliang Tang, and Dawei Yin.
\newblock Graph neural networks for social recommendation.
\newblock In {\em The world wide web conference}, pages 417--426, 2019.

\bibitem{wu2022graph}
Shiwen Wu, Fei Sun, Wentao Zhang, Xu~Xie, and Bin Cui.
\newblock Graph neural networks in recommender systems: a survey.
\newblock {\em ACM Computing Surveys}, 55(5):1--37, 2022.

\bibitem{wu2020comprehensive}
Zonghan Wu, Shirui Pan, Fengwen Chen, Guodong Long, Chengqi Zhang, and S~Yu Philip.
\newblock A comprehensive survey on graph neural networks.
\newblock {\em IEEE transactions on neural networks and learning systems}, 32(1):4--24, 2020.

\bibitem{Zhou2021OptimizingME}
Ao~Zhou, Jianlei Yang, Yeqi Gao, Tong Qiao, Yingjie Qi, Xiaoyi Wang, Yunli Chen, Pengcheng Dai, Weisheng Zhao, and Chunming Hu.
\newblock Optimizing memory efficiency of graph neural networks on edge computing platforms.
\newblock {\em ArXiv}, abs/2104.03058, 2021.

\bibitem{Kanerva2009HyperdimensionalCA}
Pentti Kanerva.
\newblock Hyperdimensional computing: An introduction to computing in distributed representation with high-dimensional random vectors.
\newblock {\em Cognitive Computation}, 1:139--159, 2009.

\bibitem{imani2019quanthd}
Mohsen Imani, Samuel Bosch, Sohum Datta, Sharadhi Ramakrishna, Sahand Salamat, Jan~M Rabaey, and Tajana Rosing.
\newblock Quanthd: A quantization framework for hyperdimensional computing.
\newblock {\em IEEE Transactions on Computer-Aided Design of Integrated Circuits and Systems}, 39(10):2268--2278, 2019.

\bibitem{khaleghi2022generic}
Behnam Khaleghi, Jaeyoung Kang, Hanyang Xu, Justin Morris, and Tajana Rosing.
\newblock Generic: highly efficient learning engine on edge using hyperdimensional computing.
\newblock In {\em Proceedings of the 59th ACM/IEEE Design Automation Conference}, pages 1117--1122, 2022.

\bibitem{kovalev2022vector}
Alexey~K Kovalev, Makhmud Shaban, Evgeny Osipov, and Aleksandr~I Panov.
\newblock Vector semiotic model for visual question answering.
\newblock {\em Cognitive Systems Research}, 71:52--63, 2022.

\bibitem{kim2020geniehd}
Yeseong Kim, Mohsen Imani, Niema Moshiri, and Tajana Rosing.
\newblock Geniehd: Efficient dna pattern matching accelerator using hyperdimensional computing.
\newblock In {\em 2020 Design, Automation \& Test in Europe Conference \& Exhibition (DATE)}, pages 115--120. IEEE, 2020.

\bibitem{li2023hypernode}
Haomin Li, Fangxin Liu, Yichi Chen, and Li~Jiang.
\newblock Hypernode: An efficient node classification framework using hyperdimensional computing.
\newblock In {\em 2023 IEEE/ACM International Conference on Computer Aided Design (ICCAD)}, pages 1--9. IEEE, 2023.

\bibitem{nunes2022graphhd}
Igor Nunes, Mike Heddes, Tony Givargis, Alexandru Nicolau, and Alex Veidenbaum.
\newblock Graphhd: Efficient graph classification using hyperdimensional computing.
\newblock In {\em 2022 Design, Automation \& Test in Europe Conference \& Exhibition (DATE)}, pages 1485--1490. IEEE, 2022.

\bibitem{zhao2021stars}
Lingxiao Zhao, Wei Jin, Leman Akoglu, and Neil Shah.
\newblock From stars to subgraphs: Uplifting any gnn with local structure awareness.
\newblock {\em arXiv preprint arXiv:2110.03753}, 2021.

\bibitem{feng2022powerful}
Jiarui Feng, Yixin Chen, Fuhai Li, Anindya Sarkar, and Muhan Zhang.
\newblock How powerful are k-hop message passing graph neural networks.
\newblock {\em Advances in Neural Information Processing Systems}, 35:4776--4790, 2022.

\bibitem{chandrasekaran2022fhdnn}
Rishikanth Chandrasekaran, Kazim Ergun, Jihyun Lee, Dhanush Nanjunda, Jaeyoung Kang, and Tajana Rosing.
\newblock Fhdnn: Communication efficient and robust federated learning for aiot networks.
\newblock In {\em Proceedings of the 59th ACM/IEEE Design Automation Conference}, pages 37--42, 2022.

\bibitem{Zhang2018AnED}
Muhan Zhang, Zhicheng Cui, Marion Neumann, and Yixin Chen.
\newblock An end-to-end deep learning architecture for graph classification.
\newblock In {\em AAAI Conference on Artificial Intelligence}, 2018.

\bibitem{gilmer2017neural}
Justin Gilmer, Samuel~S Schoenholz, Patrick~F Riley, Oriol Vinyals, and George~E Dahl.
\newblock Neural message passing for quantum chemistry.
\newblock In {\em International conference on machine learning}, pages 1263--1272. PMLR, 2017.

\bibitem{xu2018powerful}
Keyulu Xu, Weihua Hu, Jure Leskovec, and Stefanie Jegelka.
\newblock How powerful are graph neural networks?
\newblock {\em arXiv preprint arXiv:1810.00826}, 2018.

\bibitem{ding2022sketch}
Mucong Ding, Tahseen Rabbani, Bang An, Evan Wang, and Furong Huang.
\newblock Sketch-gnn: Scalable graph neural networks with sublinear training complexity.
\newblock {\em Advances in Neural Information Processing Systems}, 35:2930--2943, 2022.

\bibitem{lecun2015deep}
Yann LeCun, Yoshua Bengio, and Geoffrey Hinton.
\newblock Deep learning.
\newblock {\em nature}, 521(7553):436--444, 2015.

\bibitem{jin2020self}
Wei Jin, Tyler Derr, Haochen Liu, Yiqi Wang, Suhang Wang, Zitao Liu, and Jiliang Tang.
\newblock Self-supervised learning on graphs: Deep insights and new direction.
\newblock {\em arXiv preprint arXiv:2006.10141}, 2020.

\bibitem{wu2019simplifying}
Felix Wu, Amauri Souza, Tianyi Zhang, Christopher Fifty, Tao Yu, and Kilian Weinberger.
\newblock Simplifying graph convolutional networks.
\newblock In {\em International conference on machine learning}, pages 6861--6871. PMLR, 2019.

\bibitem{amrouch2022brain}
Hussam Amrouch, Mohsen Imani, Xun Jiao, Yiannis Aloimonos, Cornelia Fermuller, Dehao Yuan, Dongning Ma, Hamza~E Barkam, Paul~R Genssler, and Peter Sutor.
\newblock Brain-inspired hyperdimensional computing for ultra-efficient edge ai.
\newblock In {\em 2022 International Conference on Hardware/Software Codesign and System Synthesis (CODES+ ISSS)}, pages 25--34. IEEE, 2022.

\bibitem{poduval2022graphd}
Prathyush Poduval, Haleh Alimohamadi, Ali Zakeri, Farhad Imani, M~Hassan Najafi, Tony Givargis, and Mohsen Imani.
\newblock Graphd: Graph-based hyperdimensional memorization for brain-like cognitive learning.
\newblock {\em Frontiers in Neuroscience}, 16:757125, 2022.

\bibitem{kang2022relhd}
Jaeyoung Kang, Minxuan Zhou, Abhinav Bhansali, Weihong Xu, Anthony Thomas, and Tajana Rosing.
\newblock Relhd: A graph-based learning on fefet with hyperdimensional computing.
\newblock In {\em 2022 IEEE 40th International Conference on Computer Design (ICCD)}, pages 553--560. IEEE, 2022.

\bibitem{ge2020classification}
Lulu Ge and Keshab~K Parhi.
\newblock Classification using hyperdimensional computing: A review.
\newblock {\em IEEE Circuits and Systems Magazine}, 20(2):30--47, 2020.

\bibitem{yang2023device}
Junhuan Yang, Yi~Sheng, Yuzhou Zhang, Weiwen Jiang, and Lei Yang.
\newblock On-device unsupervised image segmentation.
\newblock In {\em 2023 60th ACM/IEEE Design Automation Conference (DAC)}, pages 1--6. IEEE, 2023.

\bibitem{thomas2021theoretical}
Anthony Thomas, Sanjoy Dasgupta, and Tajana Rosing.
\newblock A theoretical perspective on hyperdimensional computing.
\newblock {\em Journal of Artificial Intelligence Research}, 72:215--249, 2021.

\bibitem{hassan2021hyper}
Eman Hassan, Yasmin Halawani, Baker Mohammad, and Hani Saleh.
\newblock Hyper-dimensional computing challenges and opportunities for ai applications.
\newblock {\em IEEE Access}, 10:97651--97664, 2021.

\bibitem{aygun2023learning}
Sercan Aygun, Mehran~Shoushtari Moghadam, M~Hassan Najafi, and Mohsen Imani.
\newblock Learning from hypervectors: A survey on hypervector encoding.
\newblock {\em arXiv preprint arXiv:2308.00685}, 2023.

\bibitem{yu2022understanding}
Tao Yu, Yichi Zhang, Zhiru Zhang, and Christopher~M De~Sa.
\newblock Understanding hyperdimensional computing for parallel single-pass learning.
\newblock {\em Advances in Neural Information Processing Systems}, 35:1157--1169, 2022.

\bibitem{rahimi2016hyperdimensional}
Abbas Rahimi, Simone Benatti, Pentti Kanerva, Luca Benini, and Jan~M Rabaey.
\newblock Hyperdimensional biosignal processing: A case study for emg-based hand gesture recognition.
\newblock In {\em 2016 IEEE International Conference on Rebooting Computing (ICRC)}, pages 1--8. IEEE, 2016.

\bibitem{imani2021revisiting}
Mohsen Imani, Zhuowen Zou, Samuel Bosch, Sanjay~Anantha Rao, Sahand Salamat, Venkatesh Kumar, Yeseong Kim, and Tajana Rosing.
\newblock Revisiting hyperdimensional learning for fpga and low-power architectures.
\newblock In {\em 2021 IEEE International Symposium on High-Performance Computer Architecture (HPCA)}, pages 221--234. IEEE, 2021.

\bibitem{brin1998anatomy}
Sergey Brin and Lawrence Page.
\newblock The anatomy of a large-scale hypertextual web search engine.
\newblock {\em Computer networks and ISDN systems}, 30(1-7):107--117, 1998.

\bibitem{imani2018hierarchical}
Mohsen Imani, Chenyu Huang, Deqian Kong, and Tajana Rosing.
\newblock Hierarchical hyperdimensional computing for energy efficient classification.
\newblock In {\em Proceedings of the 55th Annual Design Automation Conference}, pages 1--6, 2018.

\bibitem{duan2022hdlock}
Shijin Duan, Shaolei Ren, and Xiaolin Xu.
\newblock Hdlock: Exploiting privileged encoding to protect hyperdimensional computing models against ip stealing.
\newblock In {\em Proceedings of the 59th ACM/IEEE Design Automation Conference}, pages 679--684, 2022.

\bibitem{imani2017voicehd}
Mohsen Imani, Deqian Kong, Abbas Rahimi, and Tajana Rosing.
\newblock Voicehd: Hyperdimensional computing for efficient speech recognition.
\newblock In {\em 2017 IEEE international conference on rebooting computing (ICRC)}, pages 1--8. IEEE, 2017.

\bibitem{nunes2023extension}
Igor Nunes, Mike Heddes, Tony Givargis, and Alexandru Nicolau.
\newblock An extension to basis-hypervectors for learning from circular data in hyperdimensional computing.
\newblock In {\em 2023 60th ACM/IEEE Design Automation Conference (DAC)}, pages 1--6. IEEE, 2023.

\bibitem{hamilton2017inductive}
Will Hamilton, Zhitao Ying, and Jure Leskovec.
\newblock Inductive representation learning on large graphs.
\newblock {\em Advances in neural information processing systems}, 30, 2017.

\bibitem{wind2012weighted}
David~Kofoed Wind and Morten Mørup.
\newblock Link prediction in weighted networks.
\newblock In {\em 2012 IEEE International Workshop on Machine Learning for Signal Processing}, pages 1--6, 2012.

\bibitem{balcilar2021breaking}
Muhammet Balcilar, Pierre H{\'e}roux, Benoit Gauzere, Pascal Vasseur, S{\'e}bastien Adam, and Paul Honeine.
\newblock Breaking the limits of message passing graph neural networks.
\newblock In {\em International Conference on Machine Learning}, pages 599--608. PMLR, 2021.

\bibitem{corso2020principal}
Gabriele Corso, Luca Cavalleri, Dominique Beaini, Pietro Li{\`o}, and Petar Veli{\v{c}}kovi{\'c}.
\newblock Principal neighbourhood aggregation for graph nets.
\newblock {\em Advances in Neural Information Processing Systems}, 33:13260--13271, 2020.

\bibitem{velickovic2017graph}
Petar Velickovic, Guillem Cucurull, Arantxa Casanova, Adriana Romero, Pietro Lio, Yoshua Bengio, et~al.
\newblock Graph attention networks.
\newblock {\em stat}, 1050(20):10--48550, 2017.

\bibitem{kleyko2022survey}
Denis Kleyko, Dmitri~A Rachkovskij, Evgeny Osipov, and Abbas Rahimi.
\newblock A survey on hyperdimensional computing aka vector symbolic architectures, part i: Models and data transformations.
\newblock {\em ACM Computing Surveys}, 55(6):1--40, 2022.

\bibitem{han2015deep}
Song Han, Huizi Mao, and William~J Dally.
\newblock Deep compression: Compressing deep neural networks with pruning, trained quantization and huffman coding.
\newblock {\em arXiv preprint arXiv:1510.00149}, 2015.

\bibitem{ijcai2020p181}
Yiqing Xie, Sha Li, Carl Yang, Raymond Chi-Wing Wong, and Jiawei Han.
\newblock When do gnns work: Understanding and improving neighborhood aggregation.
\newblock In Christian Bessiere, editor, {\em Proceedings of the Twenty-Ninth International Joint Conference on Artificial Intelligence, {IJCAI-20}}, pages 1303--1309. International Joint Conferences on Artificial Intelligence Organization, 7 2020.
\newblock Main track.

\bibitem{nikolentzos2019message}
Giannis Nikolentzos, Antoine J.~P. Tixier, and Michalis Vazirgiannis.
\newblock Message passing attention networks for document understanding, 2019.

\bibitem{kim2023efficient}
Jiseung Kim, Hyunsei Lee, Mohsen Imani, and Yeseong Kim.
\newblock Efficient hyperdimensional learning with trainable, quantizable, and holistic data representation.
\newblock In {\em 2023 Design, Automation \& Test in Europe Conference \& Exhibition (DATE)}, pages 1--6. IEEE, 2023.

\bibitem{Yan_2023}
Zhanglu Yan, Shida Wang, Kaiwen Tang, and Weng-Fai Wong.
\newblock {\em Efficient Hyperdimensional Computing}, page 141–155.
\newblock Springer Nature Switzerland, 2023.

\bibitem{kainen2020quasiorthogonal}
Paul~C Kainen and V{\v{e}}ra K{\r{u}}rkov{\'a}.
\newblock Quasiorthogonal dimension.
\newblock In {\em Beyond traditional probabilistic data processing techniques: Interval, fuzzy etc. Methods and their applications}, pages 615--629. Springer, 2020.

\bibitem{morris2020tudataset}
Christopher Morris, Nils~M Kriege, Franka Bause, Kristian Kersting, Petra Mutzel, and Marion Neumann.
\newblock Tudataset: A collection of benchmark datasets for learning with graphs.
\newblock {\em arXiv preprint arXiv:2007.08663}, 2020.

\bibitem{torchhd}
Mike Heddes, Igor Nunes, Pere Vergés, Denis Kleyko, Danny Abraham, Tony Givargis, Alexandru Nicolau, and Alex Veidenbaum.
\newblock Torchhd: An open source python library to support research on hyperdimensional computing and vector symbolic architectures.
\newblock {\em Journal of Machine Learning Research}, 24(255):1--10, 2023.

\bibitem{dataset-letter}
Kaspar Riesen and Horst Bunke.
\newblock Iam graph database repository for graph based pattern recognition and machine learning.
\newblock In {\em Structural, Syntactic, and Statistical Pattern Recognition: Joint IAPR International Workshop, SSPR \& SPR 2008, Orlando, USA, December 4-6, 2008. Proceedings}, pages 287--297. Springer, 2008.

\bibitem{dataset-bzr}
Jeffrey~J Sutherland, Lee~A O'brien, and Donald~F Weaver.
\newblock Spline-fitting with a genetic algorithm: A method for developing classification structure- activity relationships.
\newblock {\em Journal of chemical information and computer sciences}, 43(6):1906--1915, 2003.

\bibitem{dataset:synthie}
Aasa Feragen, Niklas Kasenburg, Jens Petersen, Marleen de~Bruijne, and Karsten Borgwardt.
\newblock Scalable kernels for graphs with continuous attributes.
\newblock {\em Advances in neural information processing systems}, 26, 2013.

\bibitem{dataset:proteins}
Paul~D Dobson and Andrew~J Doig.
\newblock Distinguishing enzyme structures from non-enzymes without alignments.
\newblock {\em Journal of molecular biology}, 330(4):771--783, 2003.

\bibitem{leman1968reduction}
AA~Leman and Boris Weisfeiler.
\newblock A reduction of a graph to a canonical form and an algebra arising during this reduction.
\newblock {\em Nauchno-Technicheskaya Informatsiya}, 2(9):12--16, 1968.

\bibitem{Fey/Lenssen/2019}
Matthias Fey and Jan~E. Lenssen.
\newblock Fast graph representation learning with {PyTorch Geometric}.
\newblock In {\em ICLR Workshop on Representation Learning on Graphs and Manifolds}, 2019.

\bibitem{bric}
Mohsen Imani, Justin Morris, John Messerly, Helen Shu, Yaobang Deng, and Tajana Rosing.
\newblock Bric: Locality-based encoding for energy-efficient brain-inspired hyperdimensional computing.
\newblock In {\em 2019 56th ACM/IEEE Design Automation Conference (DAC)}, pages 1--6, 2019.

\bibitem{pearson}
Zoran {\v{S}}verko, Miroslav Vranki{\'c}, Sa{\v{s}}a Vlahini{\'c}, and Peter Rogelj.
\newblock Complex pearson correlation coefficient for eeg connectivity analysis.
\newblock {\em Sensors}, 22(4):1477, 2022.

\bibitem{tsne}
Laurens van~der Maaten and Geoffrey Hinton.
\newblock Visualizing data using t-sne.
\newblock {\em Journal of Machine Learning Research}, 9(86):2579--2605, 2008.

\end{thebibliography}


\end{document}